\documentclass[letterpaper, 10 pt, conference]{ieeeconf}  

\IEEEoverridecommandlockouts                              

\usepackage{amsthm}
\usepackage{graphics} 
\usepackage{epsfig} 
\usepackage{times} 
\usepackage{amssymb}  
\usepackage{bm}
\usepackage{xcolor,soul}
\usepackage{cite}

\usepackage{ulem} 
\usepackage{hyperref}

\usepackage{float}
\usepackage{booktabs}
\usepackage{multirow}
\usepackage{makecell}
\usepackage{enumerate}
\usepackage{leftindex}
\usepackage{mathtools}
\usepackage{mathrsfs}
\usepackage[utf8]{inputenc}
\usepackage[caption=false,font=footnotesize]{subfig}
\usepackage{graphicx}
\usepackage{pifont}
\usepackage{soul}
\usepackage{bbding}

\newcounter{RNum}
\renewcommand{\theRNum}{\arabic{RNum}}

\theoremstyle{plain}

\theoremstyle{definition}

\newcommand{\Remark}{\noindent\textbf{Remark}~\refstepcounter{RNum}\textbf{\theRNum}: }
\newcommand{\NoOne}[1]{\textcolor{red}{#1}}
\newcommand{\NoTwo}[1]{\textcolor{green}{#1}}
\newcommand{\NoThree}[1]{\textcolor{blue}{#1}}

\newcommand{\re}{\mathbb{R}}

\soulregister\NoOne7
\soulregister\NoTwo7
\soulregister\NoThree7
\soulregister\Remark7

\title{\LARGE \bf
Local Reactive Control for Mobile Manipulators with\\ Whole-Body Safety in Complex Environments
}

\author{Chunxin Zheng, Yulin Li, Zhiyuan Song, Zhihai Bi, Jinni Zhou, Boyu Zhou, and Jun Ma 
 \thanks{Chunxin Zheng, Yulin, Li, Zhiyuan Song, Zhihai Bi, Jinni Zhou, and Jun Ma are with The Hong Kong University of Science and Technology, Guangzhou, China (e-mail: jun.ma@ust.hk).}
\thanks{Boyu Zhou is with the School of Artificial Intelligence, Sun Yat-sen University, Zhuhai, China (e-mail: zhouby23@mail.sysu.edu.cn)}
\thanks{Chunxin Zheng and Yulin Li contributed equally to this work.}
}%

\begin{document}

\maketitle
\thispagestyle{empty}
\pagestyle{empty}


\begin{abstract}
Mobile manipulators typically encounter significant challenges in navigating narrow, cluttered environments due to their high-dimensional state spaces and complex kinematics. While reactive methods excel in dynamic settings, they struggle to efficiently incorporate complex, coupled constraints across the entire state space. 
In this work, we present a novel local reactive controller that reformulates the time-domain single-step problem into a multi-step optimization problem in the spatial domain, leveraging the propagation of a serial kinematic chain. 
This transformation facilitates the formulation of customized, decoupled link-specific constraints, which is further solved efficiently with augmented Lagrangian differential dynamic programming (AL-DDP). Our approach naturally absorbs spatial kinematic propagation in the forward pass and processes all link-specific constraints simultaneously during the backward pass, enhancing both constraint management and computational efficiency. 
Notably, in this framework,
we formulate collision avoidance constraints for each link using accurate geometric models with extracted free regions, and this improves the maneuverability of the mobile manipulator in narrow, cluttered spaces. Experimental results showcase significant improvements in safety, efficiency, and task completion rates. These findings
underscore the robustness of the proposed method, particularly in narrow, cluttered environments where conventional approaches could falter.
The open-source project can be found at \url{https://github.com/Chunx1nZHENG/MM-with-Whole-Body-Safety-Release.git}.
\end{abstract}

\section{Introduction} \label{sec:intro}
Mobile manipulators, renowned for their versatility and extensive range of motion, are widely deployed in construction and industrial settings \cite{pankert2020perceptive}. 
Enhancing the ability of mobile manipulators to navigate safely and efficiently in unknown and cluttered environments has become a crucial research direction, driven by the demands of numerous real-world applications. However, this complex task is confronted with two significant challanges: the efficient control of mobile manipulators with high degrees of freedom and the precise modeling of collision avoidance constraints in dynamic, unstructured environments.

The first challenge lies in efficiently controlling mobile manipulators with a high degree of freedom. 
These systems typically integrate the mobile base with the multi-link manipulator, resulting in high-dimensional state spaces and intricate kinematics, which pose considerable control challenges within the robotics domain. 
Traditional methods for trajectory planning, such as sampling-based approaches  \cite{li2016asymptotically} and optimization-based techniques  \cite{kalakrishnan2011stomp}, can effectively generate global or local optimal trajectories for manipulation or navigation tasks, which can then be tracked directly with a low-level blind controller. 
These methods, while effective in generating optimal trajectories for static environments, often struggle to adapt to dynamic changes and unexpected obstacles in real time. The computational burden of solving optimization problems over a receding horizon can lead to significant latency, particularly in high-dimensional state spaces or when dealing with complex, nonlinear dynamics. These limitations underscore the need for more agile and computationally efficient approaches. On the other hand, local reactive controllers address these challenges by providing rapid, on-the-fly responses to environmental changes. By integrating real-time constraints and employing efficient algorithms, these controllers offer a promising solution for navigating complex, dynamic environments. However, the desire to incorporate precise, complex constraints for improved performance conflicts with the need for real-time computation, especially in high-dimensional systems. This issue is particularly acute in scenarios where state variables are tightly coupled, such as in mobile manipulators. Consequently, current implementations of local reactive controllers often prioritize computational speed over model complexity, leading to simplified problem formulations.

On the other hand, for accurate modeling of collision avoidance constraints for mobile manipulators in dynamic, unstructured environments, the complexity stems from the need to simultaneously consider the robot's expansive workspace, varying configurations, and dynamic obstacles. 
To address this challenge, the most common approach involves modeling collision avoidance constraints based on the distance between the robots and potential obstacles.
Given the intricate geometry of mobile manipulators and the need to simplify distance calculations, robots are generally represented as a union of spheres \cite{pankert2020perceptive} or as a collection of points \cite{haviland2021NEO}. This abstraction helps streamline the collision avoidance process while still providing a practical approximation of the robot's shape and spatial requirements.
However, these methods for describing the shapes of robots might not be precise, which potentially hinder the efficiency of collision checking. Also, this can lead to a restricted motion space and produce overly conservative control outputs. As a result, accurately representing the robot's shape and efficiently formulating safety constraints are crucial for achieving effective collision avoidance in narrow, cluttered environments.

To overcome the constraint coupling issue in local reactive controllers, we introduce a novel multi-step spatial control formulation. This approach enables the integration of complex constraints without compromising computational efficiency. Additionally, we propose an innovative formulation of collision avoidance constraints for mobile manipulators that defines the potential motion space. This formulation leverages advanced geometric representations to characterize the robot's shape more accurately, and this results in a less conservative and more efficient navigation strategy.
The key contributions of our work are:

\begin{itemize}
    \item We reconstruct the time-domain single-step control problem as a multi-step control problem in the spatial domain, following the propagation of the serial kinematic chain.
    It effectively decouples the constraints associated with each link, which facilitates the construction of customized constraints and objectives for each link individually, and this avoids the faltering of performance caused by constraint coupling.
     \item We propose an innovative formulation of collision avoidance constraints for multi-rigid-body robots by integrating precise link-specific geometric models with dynamically extracted free regions around each link. This approach significantly enhances the robot's maneuverability in narrow, cluttered environments.
      \item We solve the resulting spatial optimization problem by employing augmented Lagrangian differential dynamic programming (AL-DDP), which naturally integrates spatial kinematics and decoupled link-specific constraints. This significantly enhances computational efficiency toward optimization tasks for multi-body robotic systems, while implicitly and efficiently addressing the collision avoidance constraints in cluttered environments.
    \item Our experimental results demonstrate substantial enhancements in terms of safety, efficiency, and task completion rates within highly constrained spaces. Remarkably, the proposed approach provides new perspectives and insights for the control of high-dimensional robot systems while ensuring whole-body safety.
\end{itemize}

\section{Related Works}
\subsection{Holistic Control}
Holistic control of mobile manipulators has gained significant attention due to its ability to fully utilize all degrees of freedom in performing complex tasks. These integrated control strategies can be broadly categorized into two main approaches: trajectory planning with low-level blind control, and reactive methods.
In the first category, sampling-based methods are among the most common approaches for trajectory planning in high-dimensional spaces. However, these methods often require a large number of samples to satisfy optimality conditions in high-dimensional spaces, resulting in an extremely high computational burden. Additionally, ensuring kinematic and dynamic feasibility often requires extra computational effort \cite{7759547}. In terms of safety, these methods rely on collision checking of sampled points \cite{8998369,9340782}, making it challenging to balance between precise collision detection and computational time.
On the other hand, trajectory optimization algorithms, such as CHOMP \cite{ratliff2009chomp} and TrajOpt \cite{schulman2013finding}, can effectively incorporate complex constraints and converge to good optimal solutions when provided with a reasonable initial guess. However, these methods struggle to guarantee real-time performance in high-dimensional problems.

Given the limitations of these methods in dynamic environments, researchers have turned to the development of reactive controllers. These reactive methods directly incorporate environmental data into the closed-loop control process. They can be further divided into finite-horizon and single-step approaches. Finite-horizon methods, such as Model Predictive Control (MPC) \cite{pankert2020perceptive}, offer predictive capabilities by optimizing over a future time window. While this approach generates smoother trajectories by considering future states, extending the time horizon increases computational costs and can compromise real-time performance.
In contrast, single-step reactive controllers, often formulated as quadratic programming (QP) problems \cite{haviland2022holistic,9815144}, prioritize real-time performance. These methods can achieve smoother robot motions and faster computation times. However,
it is worth noting that most existing approaches, including the aforementioned reactive controllers, typically operate in the joint angle space, using joint positions and velocities as the primary state representation. While this joint-space formulation is intuitive and directly relates to the robot's actuators, it can lead to difficulties in expressing and enforcing link-specific constraints, especially in the context of obstacle avoidance and task-space control for mobile manipulators, making it challenging to incorporate complex constraints while ensuring real-time performance.

\subsection{Collision Avoidance}
Collision avoidance remains a significant challenge in robotics, primarily due to the complexities involved in accurately modeling obstacle avoidance constraints. Sampling-based approaches often utilize occupancy maps, such as OctoMap \cite{hornung13auro}, which divide space into occupied and free voxels. While effective for collision checking at sampled nodes, these methods require numerous collision queries, potentially compromising real-time performance.
Optimization-based methods offer an alternative approach by incorporating collision avoidance as nonlinear constraints within nonlinear programming (NLP), eliminating the need for explicit collision queries. These methods typically construct distance-related constraints to ensure obstacle avoidance. The Euclidean signed distance field (ESDF) is a popular technique for formulating such constraints \cite{han2019fiesta,wu2024real}, providing crucial distance and gradient information. However, ESDF map construction introduces computational complexity and high memory usage and poses challenges in dynamic environments.

To address these limitations, space decomposition methods have emerged as promising alternatives \cite{liu2017planning}. Building upon this concept, some methods construct collision avoidance constraints by limiting the distance between the robot and the planes defining safe regions, often using simplified robot shapes \cite{spahn2021coupled}. This approach offers a balance between computational efficiency and effective obstacle avoidance, particularly in complex and dynamic environments.
However, these simplified methods for describing the shape of robots reduce their feasible motion space. In contrast, a series of methods precisely model the robot's geometry and explore the geometric relationships between the robot and free regions or obstacles to formulate collision avoidance constraints \cite{li2024geometry,li2024collision,li2024frtreeplannerrobotnavigation,tracy2023differentiable}. Nevertheless, the complexity of multi-rigid-body robots' shapes and the coupling between links pose significant challenges in applying these methods to mobile manipulators.

\section{{Methodology}}
 \begin{figure}[t]	
	\centering
	\includegraphics[width=\linewidth]{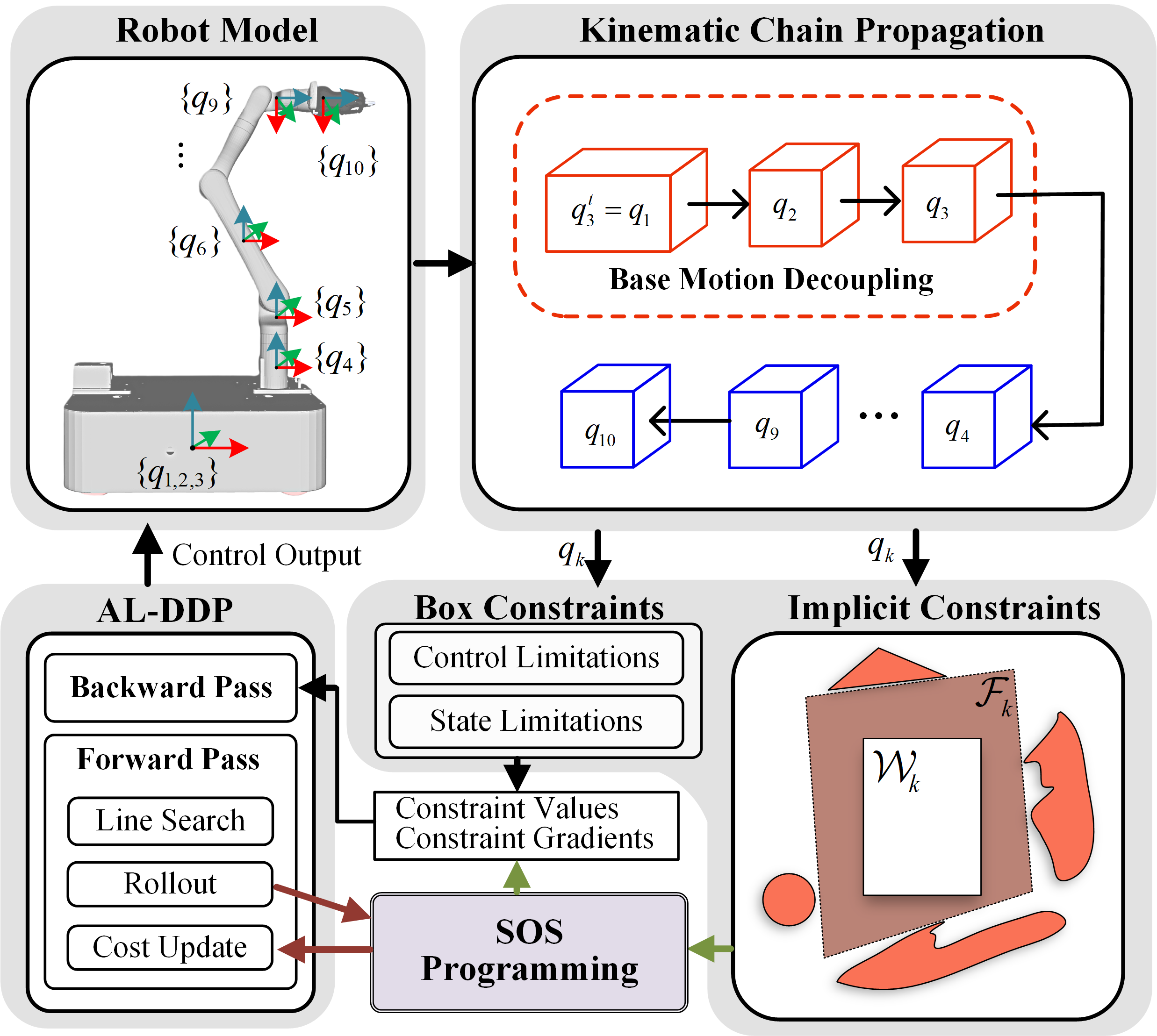}
	\setlength{\abovecaptionskip}{-0pt} 
	\caption
	{Overview of the proposed local reactive controller design approach.  We formulate the single-step reactive control problem as a finite-horizon spatial trajectory optimization problem along the kinematic chain. Using the robot states at $t$, we construct the kinematic chain propagation function to formulate the link-specific constraints. The collision avoidance constraints are formulated as implicit constraints, with the values and gradients determined using SOS programming. The spatial trajectory optimization problem is solved using the AL-DDP algorithm, where all link-specific constraints are integrated using the augmented Lagrangian.}
	\label{fig:illu_decoup}
\end{figure}

This work aims to develop a safety-critical local reactive controller for mobile manipulators based on spatial propagation along their kinematic chain.

To ensure whole-body motion safety for the articulated system, where each link's state is influenced by all preceding degrees of freedom, we formulate an optimization-based holistic control problem that expands along the kinematic chain's propagation. Specifically, the configuration of adjacent links propagates spatially along the kinematic chain through their connected joint. Fig. \ref{fig:illu_decoup} illustrates the overview of the controller design pipeline, which includes the robot modeling, constraint formulation, and problem solving process. 

\subsection{Problem Formulation}
Compared to the time-domain single-step controller, which uses joint angles as decision variables, we define the state for each link $i$ on the kinematic chain as $\boldsymbol{q}_i = [\boldsymbol{p}_i,\boldsymbol{r}_i] \in \mathbb{R}^7$, where $\boldsymbol{p}_i$ represents the 3D Euclidean position and $\boldsymbol{r}_i$ represents the orientation quaternions. The corresponding control input is denoted by $u_i \in \mathbb{R}$.

Inspired by time-domain trajectory optimization, our control problem is framed as multi-step spatial trajectory optimization over a spatial horizon $\Bar{N}$ along the entire kinematic chain. We define the spatial trajectory and control sequence as $\mathcal{Q} = \left\{{\boldsymbol{q}^{base}_x, \boldsymbol{q}^{base}_y, \boldsymbol{q}^{base}_{\phi}, \boldsymbol{q}^{arm}_1, \ldots, \boldsymbol{q}^{arm}_N}\right\}$ 
and $\boldsymbol{U} = \left\{v_x^{base}, v_y^{base}, \dot{\phi}^{base}, u_1^{arm}, \ldots, u_N^{arm}\right\}$. For convenience, we use $k$ to denote the corresponding index in $\mathcal{Q}$ and $\boldsymbol{U}$. When $k \in \{1, 2, 3\}$, $\boldsymbol{q}_k$ and ${u}_k$  represent the elements corresponding to the mobile base. 
When $k \in \{4, \ldots, \Bar{N}\}$, 
they represent the corresponding state and output of the articulated arm. The original points of the virtual links $\boldsymbol{q}_1$, $\boldsymbol{q}_2$, and $\boldsymbol{q}_3$ are located at the center of the mobile base. The coordinates of the other links are distributed along the kinematic chain of the manipulator, positioned at the center of each link. Then, at the current time step $t$ with the observation $\mathcal{Q}_t$, we introduce the following spatial trajectory optimization problem that solves for the optimal control sequence $U_t$, such that the spatial state at the next time step $\mathcal{Q}_{t+1}$ remains safe while tracking the desired target. Unlike typical trajectory optimization problems that explore optimal control sequence over the time horizon, our proposed method seeks to find the optimal control for each joint along the kinematic chain:

\begin{align}\label{eqn:to}
    \displaystyle  \operatorname*{minimize}_{\mathcal{Q}, \boldsymbol{U}}\quad 
    & \phi(\boldsymbol{q}_{\Bar{N}+1})+\sum_{k=1}^{\Bar{N}}
      J_k\big(\boldsymbol{q}_{k},{u}_{k}\big)\\
    \operatorname*{subject\ to}\quad \,\, 
    &\boldsymbol{q}^{t+1}_{k+1}=f_k\big(\boldsymbol{q}^{t}_{k+1},\boldsymbol{q}^{t+1}_{k},u_k\big) \tag{\ref{eqn:to}{a}} \\
    & h_{k}(\boldsymbol{q}^{t+1}_{k},{u}_{k}) = 0,\tag{\ref{eqn:to}{b}}\\ 
    &g_{k}(\boldsymbol{q}^{t+1}_{k},{u}_{k}) \leq 0,\yesnumber \tag{\ref{eqn:to}{c}} \\
        &{{\mathcal{W}}}_k \subseteq {{\mathcal{F}}}_k, \tag{\ref{eqn:to}{d}}\\ 
    & \quad \,\,\,\, k = 1,2,\ldots,\Bar{N}. \notag
\end{align}

In this problem, $\boldsymbol{q}^t_k$ denotes the state of the $k\text{th}$ link known from current observation, while $u_k$ and $\boldsymbol{q}^{t+1}_k$ represent the current control and the optimal state at the next step to be determined.
$J_k\big({q}_{k},{u}_{k}\big)$ and $\phi$ denote the intermediate and terminal costs, respectively. $f_k$ is the spatial kinetic propagation function between adjacent links. $h_{k}(\boldsymbol{q}_{k},{u}_{k})$ and $g_{k}(\boldsymbol{q}_{k},{u}_{k})$ are the general equality and inequality state input constraints corresponding to each link. Importantly, the constraint ${{\mathcal{W}}}_k \subseteq {{\mathcal{F}}}_k$ represents that the $k$th link is contained in the corresponding free region, thus guaranteeing that geometry-aware and collision-free movements encompass the entire robot.

\subsection{{Cost Function}}

In this optimization problem, the cost function in (\ref{eqn:to}) is designed as the quadratic form of specific user-defined error terms. Specifically, the terminal cost $\phi$ represents the cost of the end-effector tracking and the intermediate cost  $J_k$ integrates the control effort minimization for each joint and state tracking error over the $\Bar{N}$ robot links:
\begin{align*}
& \phi(\boldsymbol{q}_{\Bar{N}+1}) = \boldsymbol{e}_e^T \boldsymbol{Q}_\phi\boldsymbol{e}_e, \\
&  J_k\big(\boldsymbol{q}_{k},{u}_{k}\big) = \boldsymbol{e}^T_k\boldsymbol{Q}_k\boldsymbol{e}_k + u^T_k{R}_ku_k,
\end{align*}
where $\boldsymbol{Q}_\phi, \boldsymbol{Q}_k \in \re^{6 \times 6}$ are diagonal, positive semidefinite, weight matrices, ${R}_k \in \re$ is a positive weight parameter.
The pose tracking error for each link and end-effector can be represented as
\begin{align*}
&\boldsymbol{e}_k = [{\boldsymbol{e}_{p}^{k}} , {\boldsymbol{e}_{r}^{k}}] \in \re^6, \\
&\quad \,\,\, k = 1, 2, \ldots,\Bar{N}, \notag \\
&\boldsymbol{e}_e = [{\boldsymbol{e}_{p}^{\Bar{N}+1}} , {\boldsymbol{e}_{r}^{\Bar{N}+1}}] \in \re^6,
\end{align*}
where the translational error for each link and end-effector are represented as $\boldsymbol{e}_p^{k}$ and $\boldsymbol{e}_p^{\Bar{N}+1}$, respectively, and can be calculated using Euclidean distance between two position vectors. To compute the quaternion rotation error $\boldsymbol{e}_r \in \re^3$ between the current quaternion $\boldsymbol{r} = [r_x, r_y, r_z, r_w] \in \re^4$ and the desired quaternion $\hat{\boldsymbol{r}} = [\hat{r}_x, \hat{r}_y, \hat{r}_z, \hat{r}_w] \in \re^4$ of each link and end-effector, we formulate $\boldsymbol{e}_r$ as follows \cite{siciliano2009robotics}:
\begin{align*}
&\boldsymbol{e}_r^{k}(\boldsymbol{r}_k, \hat{\boldsymbol{r}_k})= r_w^k \cdot\left[\hat{r}_x^k, \hat{r}_y^k, \hat{r}_z^k\right]^T-\hat{r}_w^k \cdot\left[r_x^k, r_y^k, r_z^k\right]^T \\
&\quad\quad\quad\quad\quad\,  +\left[\hat{r}_x^k , \hat{r}_y^k, \hat{r}_z^k\right]^T \times\left[r_x^k, r_y^k, r_z^k\right]^T, \\
&\quad\quad k = 1, 2,  \ldots, \Bar{N}+1. \notag
\end{align*}

\Remark{
By leveraging the new features of our proposed spatial trajectory optimization framework, we can effortlessly design customized cost functions for each link, allowing for selective tracking of specific parts of the robot's structure. For instance, if the task is primarily focused on tracking the predefined trajectory of the end-effector, $\boldsymbol{Q}_k$ can be configured as a zero matrix, which ensures that the objective function is solely influenced by the tracking error of the end-effector.
}

\begin{figure}[t]	
	\centering
	\includegraphics[width=0.96\linewidth]{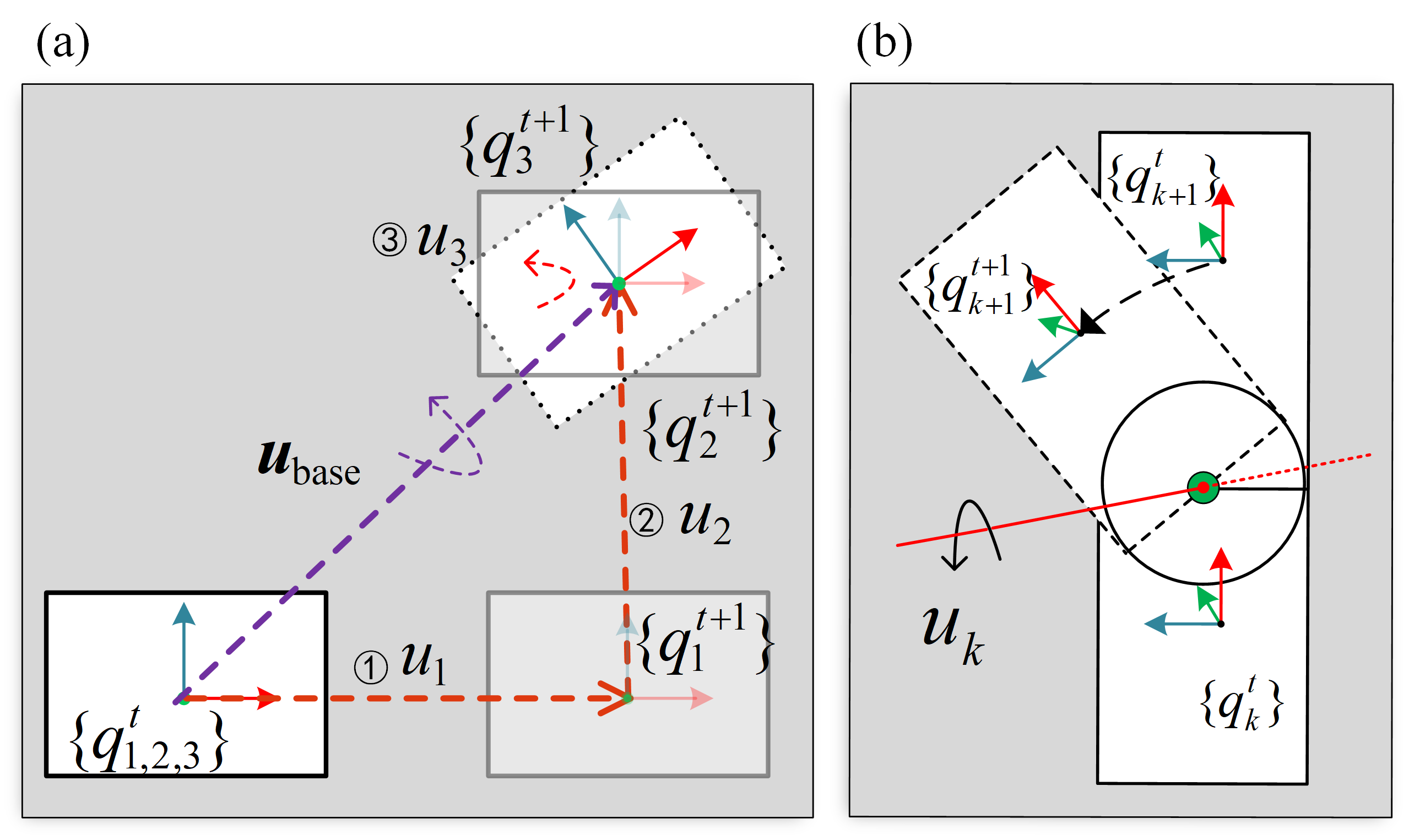}
	\setlength{\abovecaptionskip}{-0pt} 
	\caption
	{Illustration of spatial kinematic propagation of base and manipulator's links. (a). The motion process (the purple lines) generated by $\boldsymbol{u}_{base}$ of the mobile base can be decoupled into three independent stages (the dashed red lines). (b). Motion process of the manipulator's link. The dashed line link, located at $\{q_{k+1}^{t+1}\}$, is rotated from its position at $\{q_{k+1}^t\}$ along the red axis.  }
	\label{fig:illu_ab}
\end{figure}
\subsection{{Kinematic Chain Propagation}}
This section introduces the details of the kinematic chain propagation functions $f_k$ in ({\ref{eqn:to}{a}}).
 For the sake of clarity, we omit the time step superscript $t+1$ in $\boldsymbol{q}^{t+1}_{k}$. We attempt to explore how these propagation functions describe the relationship between consecutive links in the kinematic chain. They effectively map the configuration of one link to the next through the joint that connects them, which forms the foundation for the proposed spatial trajectory optimization:
\begin{subequations}\label{eqn:kin}
\begin{align}
    & \boldsymbol{q}_1 =  M(T(\boldsymbol{q}^t_3) \cdot T(u_1\cdot \Delta{t})), \\
    & \boldsymbol{q}_2 =  M(T(\boldsymbol{q}_1) \cdot T(u_2\cdot \Delta{t})), \\
    & \boldsymbol{q}_3 =  M(T(\boldsymbol{q}_2) \cdot T(u_3\cdot \Delta{t})), \\
    & \boldsymbol{q}_4 =  M(T(\boldsymbol{q}_3) \cdot T(\boldsymbol{q}_3^t,\boldsymbol{q}_4^t) \cdot T(u_4\cdot \Delta{t})), \\
    & \quad\,\,\,\vdots \notag\\
    & \boldsymbol{q}_{\Bar{N}} =  M(T(\boldsymbol{q}_{\Bar{N}-1}) \cdot T(\boldsymbol{q}_{\Bar{N}-1}^t,\boldsymbol{q}_{\Bar{N}}^t) \cdot T(u_{\Bar{N}}\cdot \Delta{t})).
\end{align}
\end{subequations}

In this context, we define $T(\cdot)$ and $T(\cdot, \cdot)$ as functions that {map different kinds of inputs to} $4 \times 4 $ transformation matrices. These functions serve three primary purposes:
\begin{itemize}
\item[$\bullet$] When the input of $T$ is a single configuration variable, such as $\boldsymbol{q}_1$, $T(\cdot)$ maps the state variable of a link to its spatial configuration in the world frame.
\item[$\bullet$] Transformation matrix generated from a control input (either rotation or translation), as shown in Fig. \ref{fig:illu_ab}.
\item[$\bullet$] In cases where $T$ involves two configuration variables, e.g.,  $T(\boldsymbol{q}_k, \boldsymbol{q}_{k+1})$, it represents the transformation from link $k$ to $k+1$.
\end{itemize}
Note that the function $M(\cdot)$ is used to map the $4 \times 4 $ transformation matrix back to a configuration vector. 

The functions {(\ref{eqn:kin}a)}-{(\ref{eqn:kin}c)} introduce the kinematic function propagation of the mobile base. $\boldsymbol{q}_3^t$ represents the current pose of the mobile base getting from the sensor. $u_1\cdot \Delta{t}$ signifies movements along the $x$-axis of the mobile base at speed $u_1$ for $\Delta t$. As shown in Fig. \ref{fig:illu_ab}(a), after moving along the $x$-axis with input $u_1$, the robot at $\boldsymbol{q}_3^t$ reaches pose $\boldsymbol{q}_1$. Similarly, we can calculate  $\boldsymbol{q}_3$ along the kinematic chain propagation \normalsize{\textcircled{\footnotesize{1}}} $\rightarrow$  \normalsize{\textcircled{\footnotesize{2}}} $\rightarrow$  \normalsize{\textcircled{\footnotesize{3}}}. The functions {(\ref{eqn:kin}d)}-{(\ref{eqn:kin}e)} and Fig. \ref{fig:illu_ab}(b) illustrate the  kinematic chain propagation of the manipulator. 
As shown in (\ref{eqn:kin}d), $T(\boldsymbol{q}_3)$ represents the pose of the base located at $\{\boldsymbol{q}_3\}$. The term $T(\boldsymbol{q}_3^t,\boldsymbol{q}_4^t)$ represents the relative pose between two links, derived from the known data at time $t$. After applying the control input $u_4$, $q_4$ at $t+1$ can be determined using  (\ref{eqn:kin}d). Along this kinematic chain propagation, we can calculate the pose of each link and $\boldsymbol{q}_{\Bar{N}}$ during the propagation process.

 \begin{figure}[t]	
	\centering
	\includegraphics[width=0.92\linewidth]{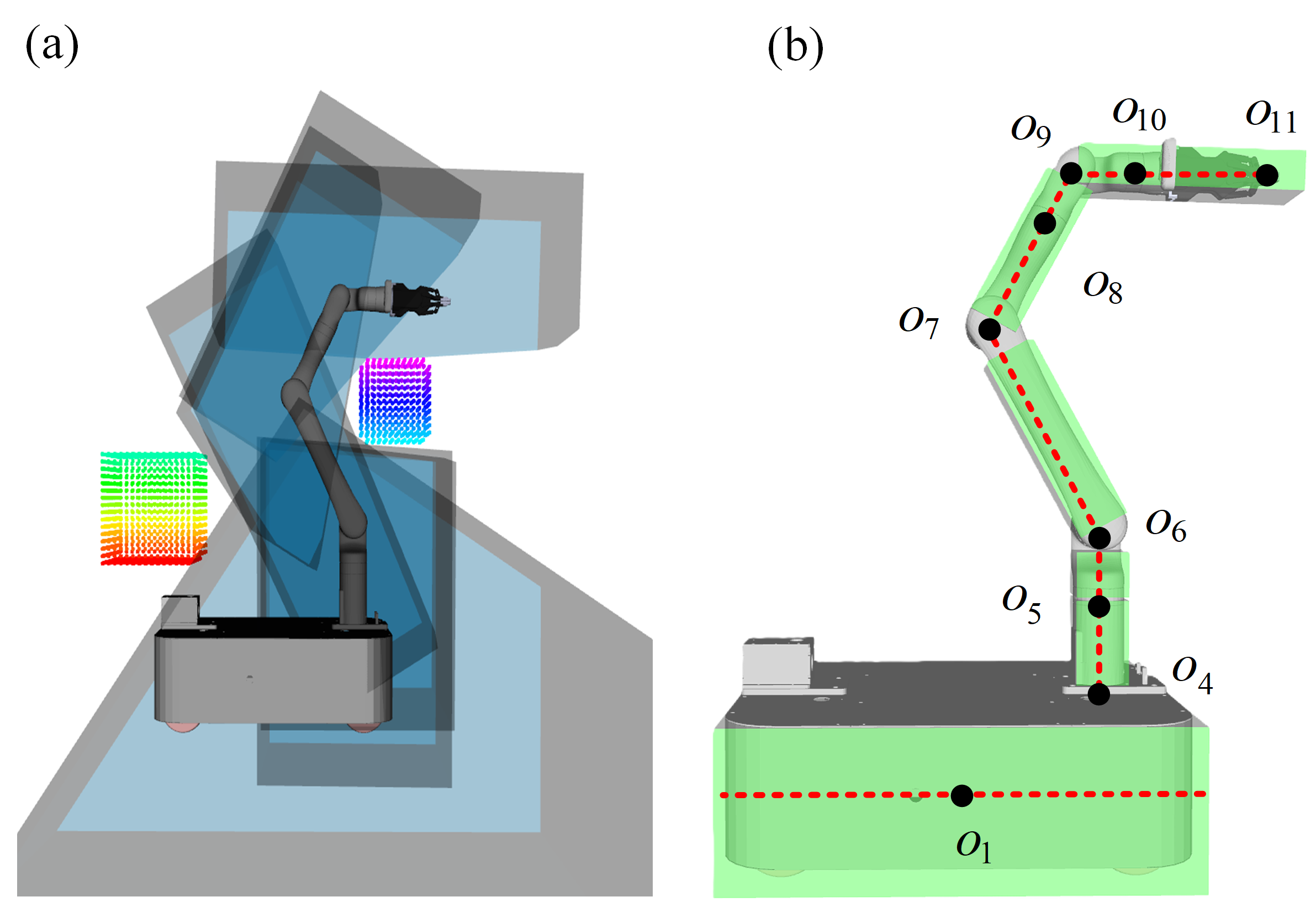}
	\setlength{\abovecaptionskip}{-0pt} 
	\caption
	{(a). The free region (depicted in blue) for each link is generated along the robot's kinematic chain. (b). The midline of each link is illustrated as red dashed lines. The mobile manipulator is described as several polytopic regions (the light green regions).}
	\label{fig:illu_space}
\end{figure}
\subsection{{Collision Avoidance }}
In \cite{li2024collision}, we formulate an SOS subproblem to calculate the minimum scaling factor of the free region to explore the geometric containment relationship between a single rigid body and the free region. Collision avoidance conditions can then be implicitly enforced by limiting the minimal scaling factor in the trajectory optimization problem. In this subsection, based on the decoupled spatial trajectory optimization framework, we extend the idea to model precise collision avoidance constraints for each link.
As depicted in Fig. \ref{fig:illu_space}(a), for every link, we generate polytopic free region along the robot's skeleton (red dashed lines in Fig. \ref{fig:illu_space}(b)) using the decomposition algorithm in \cite{liu2017planning}, denoted as ${\mathcal{F}}_k$, with $k = 1,2,  \dots, \Bar{N}$. To generate free regions with appropriate sizes and shapes to better contain the specific link and describe the surrounding movable space, we add bounding boxes when generating these free regions. Specifically, the maximum range of the base's free region is set to $0.8 \, \text{m} \times 0.8 \, \text{m}  \times 0.8 \, \text{m} $, while the maximum range of the manipulator link's free region is set to $0.4 \, \text{m}  \times 0.3 \, \text{m}  \times 0.3 \, \text{m} $. 

The polytopic free region is defined by $r_k$ linear inequalities in its own frame $\{{s_k}\}$, with $\boldsymbol{g}_k \in \mathbb{R}^{r_k}$ and $\boldsymbol{F}_k \in \mathbb{R}^{r_k \times 3}$:
\begin{equation*}
  {}^{s_k}{\mathcal{F}}_k:= \{ \boldsymbol{x}\in\re^3: \boldsymbol{g}_k-\boldsymbol{F}_k\boldsymbol{x}\ge \boldsymbol{0} \}.
\end{equation*}
The origin of $\{s_k\}$ is located at the geometric center of $\mathcal{F}_k$, with the axis aligning to the world frame $\{w\}$. Consequently, the uniformly scaled free region with scaling factor $\alpha
_k$ is:
\begin{equation*}
  {}^{s_k}{\mathcal{F}}_k(\alpha_k):= \{ \boldsymbol{x}\in\re^3: \alpha_k\boldsymbol{g}_k-\boldsymbol{F}_k\boldsymbol{x}\ge \boldsymbol{0} \}.
\end{equation*}
Subsequently, we define the space occupied by each robot link in its respective body frame $\{q_k\}$ as ${}^{q_k}{\mathcal{W}}_k$, characterized by a set of $m_k$ polynomial inequalities:
\begin{equation*}
  {}^{q_k}{\mathcal{W}}_k:= \{ \boldsymbol{x}\in\re^3: f_{1}(\boldsymbol{x})\ge0,\ldots, f_{m_k}(\boldsymbol{x})\ge0 \}.
\end{equation*}
We enforce the whole-body safety condition by limiting that each robot link is contained within its respective obstacle-free region after each step of movement. 
\begin{equation} 
  {}^{q_k}{\mathcal{W}}_k \subseteq {}^{q_k}{\mathcal{F}}_k(\alpha_k,\boldsymbol{q}_k), \quad  k = 1,2 \ldots,\Bar{N}. \label{eqn:safety_con}
\end{equation}
For convenience, we describe the containment relationship in (\ref{eqn:to}{d}) in the respective robot body frame $\{q_k\}$, such that ${}^{q_k}{\mathcal{W}}_k$ will not change, and we only need to update the free region representation upon receiving new sensor information.

Following \cite{li2024collision}, the minimum scaling factor $\alpha_k$ that satisfies (\ref{eqn:safety_con}) can be calculated from an SOS subproblem, which is further utilized to enforce safety conditions in (\ref{eqn:to}) by constraining:
\begin{equation}\label{eqn:safety_con2}
    \alpha_k \leq 1,  \quad  k = 1,2 \ldots,\Bar{N}, 
\end{equation}
Gradient information of the scaling factor $\alpha_k$ with respect to the state variables $\boldsymbol{q}_k$ is extracted from the SOS programming to further integrate with gradient-based trajectory optimizers.

 \begin{figure*}[htb]	
	\centering
        \vspace{3pt}
    \includegraphics[width=0.90\linewidth]{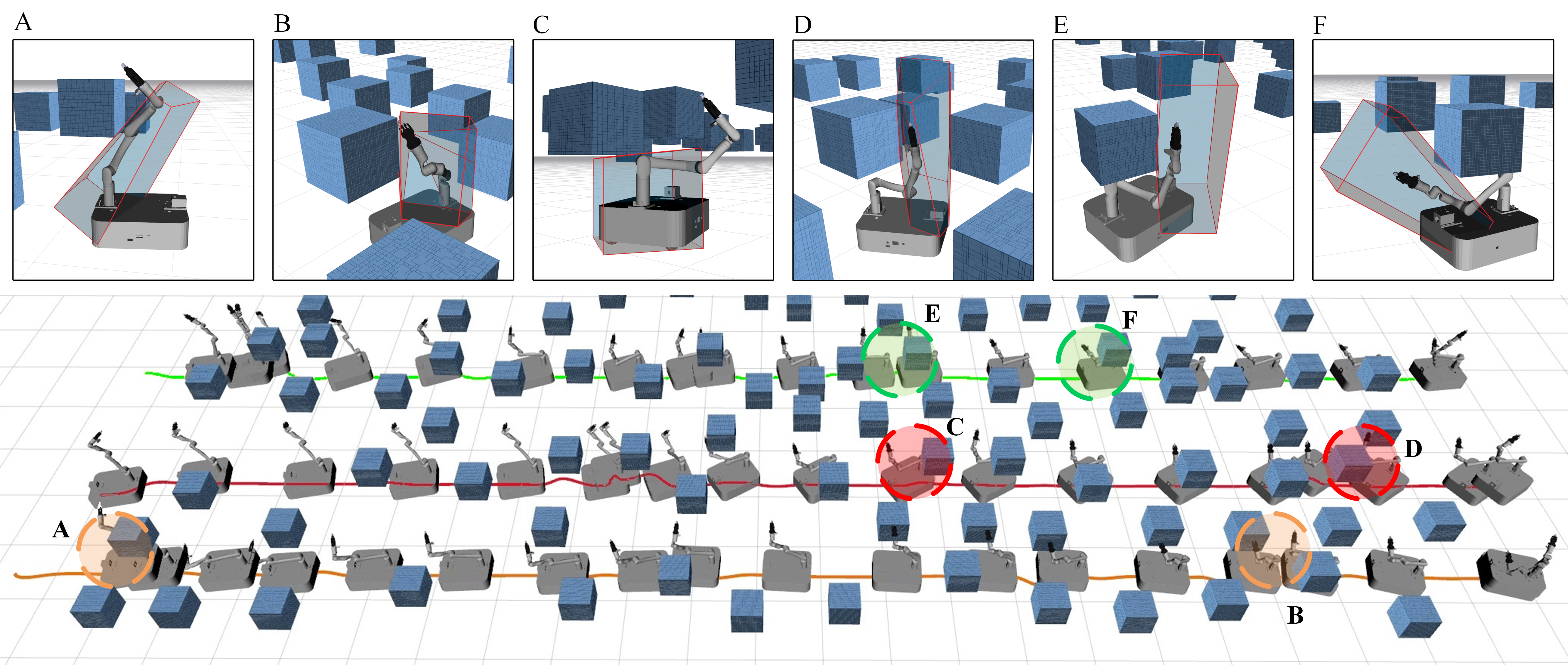}
	\caption
	{Navigation of the mobile manipulator in the random forest environment. Three trajectories of the base from different start points are represented as green, red, and orange lines, respectively. Key frames (A, B, C, D, E, F) are extracted from rviz and each key frame shows the most representative free region among all the free regions with the current robot configuration.}
	\label{fig:illu_forest}
\end{figure*}

\subsection{{Box Constraints}}
To ensure the robot's motion adheres to physical limitations, such as limits of joint velocities, mobile base speed, and joint rotation ranges, we formulate the inequality constraints ({\ref{eqn:to}{c}}) as box constraints:

\begin{align}
    & {\theta}_{k} \in [\underline{\theta},\overline{\theta}], \notag\\
    &\theta_k = \theta^t_k + u_k\Delta t, \notag \\
    &\quad\,\,\, k = 4,5 \ldots,\Bar{N},\notag \\
    & {u}_{k} \in [\underline{{u}},\overline{{u}}], \notag \\ 
    &\quad\,\,\, k = 1,2,\ldots,\Bar{N},\notag
\end{align}
where $\theta_k$ is the joint encoder data of the $k$th link. $\theta^t_k$ is the current encoder data from the manipulator.
$\underline{{\theta}}, \overline{\theta}, \underline{{u}}, \overline{{u}} \in \re$ represent the upper and lower bounds of $\theta_k$ and $u_k$, respectively. 
\subsection{{Problem Solving}}
To solve the formulated spatial trajectory optimization problem (\ref{eqn:to}) with the constraints illustrated above, we adopt Augmented Lagrangian Differential Dynamic Programming (AL-DDP). This method is particularly effective for solving optimal control problems in the time domain.

In our approach, constraints are initially augmented into the cost function. During the forward pass, AL-DDP propagates the system dynamics, which corresponds to the spatial propagation of the mobile manipulator in our case. This allows for efficient updates of link configurations based on current control inputs, effectively capturing the robot's movement through space.
The backward pass of our algorithm leverages the decoupled nature of our problem formulation. Here, we simultaneously extract all link-specific constraint values and their gradients. The extracted information is then utilized to compute optimal control commands that balance both optimality and constraint satisfaction.

By integrating our spatial formulation with AL-DDP, we efficiently address the challenges posed by the complex, high-dimensional state space of mobile manipulators. This approach enables real-time optimization while effectively managing the intricate constraints inherent to multi-body robotic systems operating in dynamic environments. The synergy between our spatial formulation and AL-DDP results in a robust method capable of rapid decision-making and precise motion control in cluttered and unpredictable settings.

\section{Results}
In this section, we validate the safety and effectiveness of our proposed approach in various scenarios, using both simulations and real-world experiments. For simulations, the method is tested on a laptop equipped with an Intel i7-13300H processor. Firstly, we evaluate the overall performance of our controller in a cluttered forest setting with a rough base reference path. Secondly, we design complex manipulation tasks to compare our approach with several baseline methods. 
Finally, we deploy our approach on an omnidirectional mobile robot equipped with a 6-DOF Kinova GEN3 manipulator. The low-level controller runs on the robot's internal computer with an Intel i5-8600 processor, while the proposed algorithm is executed on the laptop.
The spatial trajectory optimization problem is solved using ALTRO \cite{howell2019altro},  through the bi-level process interacting with the SOS subproblems to ensure safety for each link, as detailed in our previous study \cite{li2024collision}. The SOS programming problem is transformed into an SDP and solved using the conic programming solver COPT \cite{ge2022cardinal}.

\subsection{Simulations}
In this section, we evaluate the performance of our approach through two experiments. The first experiment involves a forest scenario, where the robot navigates through a random forest. The second is a multifaceted experiment that combines both navigation and manipulation tasks.

In both scenarios, we compare our proposed method with four benchmarks. Our primary focus is ensuring whole-body, collision-free movement while maintaining precise tracking of the predefined path. The benchmarks include RRT, a sampling-based method; SLQ-MPC \cite{pankert2020perceptive}, which uses the ESDF for safety constraints; Coupled-MPC \cite{spahn2021coupled}, which calculates the distance between the robot and the planes of the free region; and NEO \cite{haviland2021NEO}, a single-step QP controller.
These comparisons aim to demonstrate the effectiveness and efficiency of our method in achieving safe and precise control across various challenging environments.

\begin{table}[htb]
  \centering
  \caption{COMPARISON RESULT OF OUR METHOD WITH OTHER FOUR BENCHMARKS IN FOREST SIMULATION EXPERIMENT}
  \resizebox{0.5\textwidth}{!}{
    \begin{tabular}{l|cccc}
    \toprule
Algorithm   & \makecell{Success \\ Rate} 
                & \makecell{Path \\ Length} 
                & \makecell{Real-time \\ \,} 
                & \makecell{Reference \\ Requirement} \\
    \midrule
    RRT   & $1$ & $23.61$ & \textcolor{red}{\ding{55}} & \textcolor{red}{\ding{55}} \\
    SLQ-MPC  & $0.9$ & $28.16$ & \textcolor{green}{\ding{51}} & \textcolor{green}{\ding{51}} \\
    Coupled-MPC  & $0.6$ & $28.33$ & \textcolor{green}{\ding{51}} & \textcolor{green}{\ding{51}} \\
    NEO   & \text{\text{N/A}} & \text{\text{N/A}} & \text{\text{N/A}} & \text{\text{N/A}} \\
    Ours  & $\bm{0.9}$ & $\bm{26.24}$ & \textcolor{green}{\ding{51}} & \textcolor{green}{\ding{51}} \\
    \bottomrule
    \end{tabular}%
    }
  \label{tab:forestLabel}%
\end{table}%

\subsubsection{Cluttered Forest}In the forest simulation experiment, we aim to demonstrate the robustness and effectiveness of our method. We test it in a $20 \, \text{m} \times 10 \, \text{m} \times 3 \, \text{m}$ forest, where obstacles are randomly generated at two different densities: 0.4 $\text{obstacles}/\text{m}^2$ and 0.7 $\text{obstacles}/\text{m}^2$, as illustrated in Fig. \ref{fig:illu_forest}. Each obstacle is a $20 \, \text{cm} \times 20 \, \text{cm} \times 20 \, \text{cm}$ block. In addition, we select three start-goal pairs at the map's edge, connecting the start-goal pairs to generate global initial paths for the mobile base, providing rough and unsafe guidance. For each benchmark method, we perform 20 repetitions for each path, with the perception range set to a sphere with a radius of $3 \, \text{m}$, collecting statistics on safety, success rate and path length in Table \ref{tab:forestLabel}. It is important to note that due to the extended solving time of NEO when surrounded by numerous obstacles, NEO's data are not included in the results. Each successful arrival at the goal without any collision will be recorded as a complete task.
Typical navigation paths generated by our method are visualized in Fig. \ref{fig:illu_forest}, where we provide detailed illustrations of key frames labeled as A, B, C, D, E, F along the path.

 In the simulation, we employ the RRT method, leveraging the entire environmental information to identify collision-free paths offline. 
 However, RRT struggles to achieve real-time control, even with a limited perception range, primarily due to the high dimensionality of mobile manipulators. The SLQ-MPC and Coupled-MPC approaches both use oversized spheres to approximate the robot's shape, leading to conservative trajectories that increase travel distance. Furthermore, these methods are not well-suited for deployment in narrow environments.

Our method does not rely on a predefined environment and it considers the precise shape of each robot link. This capability motivates the robot to effectively navigate narrow environments. 
However, in certain failure cases, the accurate representation of the robot's shape may result in highly aggressive control outputs with minimal clearance. As a result, control inaccuracies and imprecise initial paths may lead to minor scratches. 

 \begin{figure}[t]	
	\centering
	\includegraphics[width=0.85\linewidth]{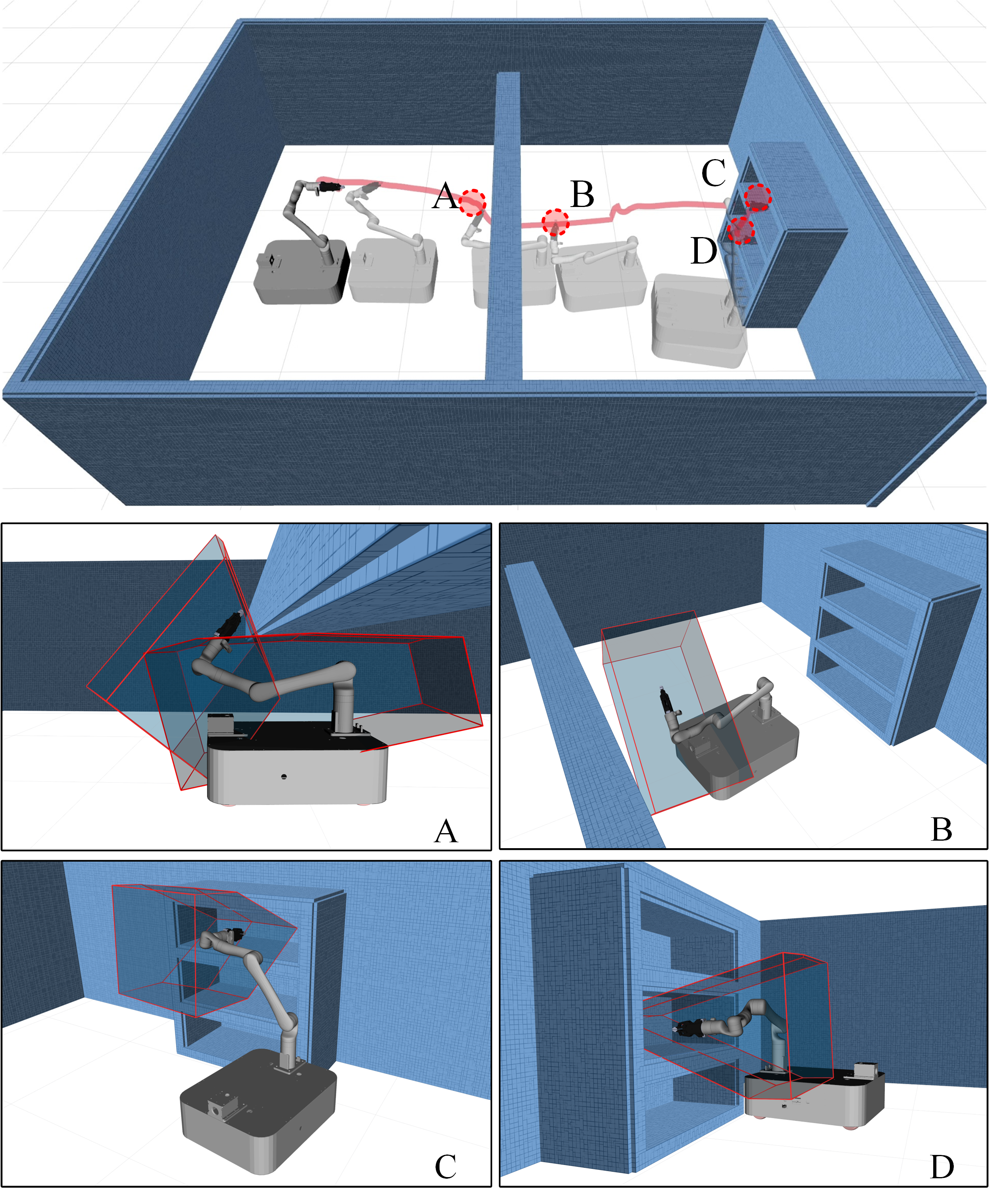}
	\setlength{\abovecaptionskip}{-0pt} 
	\caption
	{Visualization of the manipulation task in the unstructured environment. The top figure shows the whole trajectory of the manipulation task, where red circles indicate key frames (A, B, C, D). A: The robot encounters a floating bar. The mobile base continues moving forward while the manipulator adjusts to avoid the bar. B: The robot is positioned under the floating bar without any collisions. C: The end-effector is tracking the target position and orientation at the top layer of the bookcases, which has a small clearance. D: The end-effector moves to the second target located at the middle layer of the bookcases. }
	\label{fig:illu_table}
\end{figure}
\subsubsection{Multifaceted Task}
In this unstructured environment, we design two common elements: a floating bar and three-layer bookcases, as illustrated in Fig. \ref{fig:illu_table}. These elements are used to validate the robot's capabilities in locomotion, manipulation, and collision avoidance. The experiment consists of three stages. In the first stage, the robot moves from the starting point to drill through the floating bar. In the second stage, the robot tracks the target pose of the end-effector at the top layer. In the third stage, the robot moves the end-effector from the top layer to the middle layer. We generate a fixed, rough, and initially unsafe global path for the end-effector to guide the robot in completing the tasks. Additionally, we also set the same perception range as the forest experiment.

 RRT can utilize global environment information to generate safe paths offline. SLQ-MPC explicitly couples all constraint violations and gradients during the solving process, which may result in a local optimum. Meanwhile, Coupled-MPC uses a rough description of the robot, making the algorithm prone to failure in the third stage of the experiment. NEO is designed as a velocity controller, so control failures occur when the optimizer fails to find a solution that adheres to the robot's joint velocity limits.
Our method demonstrates precise and effective obstacle avoidance, resulting in the discovery of shorter paths compared to other benchmark methods. A key feature of our approach is the accurate geometric modeling of the robot, particularly the end-effector link. This enables the robot to maximize the use of its available motion space, allowing for effective maneuvers in tightly constrained environments while ensuring safety.

\begin{table}[t]
  \centering
  \caption{COMPARISON OF OUR METHOD WITH OTHER FOUR BENCHMARKS IN MULTIFACETED TASK}
  \resizebox{0.5\textwidth}{!}{%
    \begin{tabular}{l|cccc}
    \toprule
Algorithm   & \makecell{Success \\ Rate} 
                & \makecell{Path \\ Length} 
                & \makecell{Real-time \\ \,} 
                & \makecell{Reference \\ Requirement} \\
    \midrule
    RRT         & $1$          & $11.25$     & \textcolor{red}{\ding{55}} & \textcolor{red}{\ding{55}} \\
    SLQ-MPC     & $0.8$        & $5.8$       & \textcolor{green}{\ding{51}} & \textcolor{green}{\ding{51}} \\
    Coupled-MPC & $0.6$        & $7.21$      & \textcolor{green}{\ding{51}} & \textcolor{green}{\ding{51}} \\
    NEO         & $0.2$        & $6.13$      & \textcolor{green}{\ding{51}} & \textcolor{green}{\ding{51}} \\
    Ours        & $\bm{0.95}$  & $\bm{5.72}$ & \textcolor{green}{\ding{51}} & \textcolor{green}{\ding{51}} \\
    \bottomrule
    \end{tabular}%
  }
  \label{tab:shelfLabel}%
\end{table}%

 \begin{figure*}[t]	
	\centering
	\includegraphics[width=0.92\linewidth]{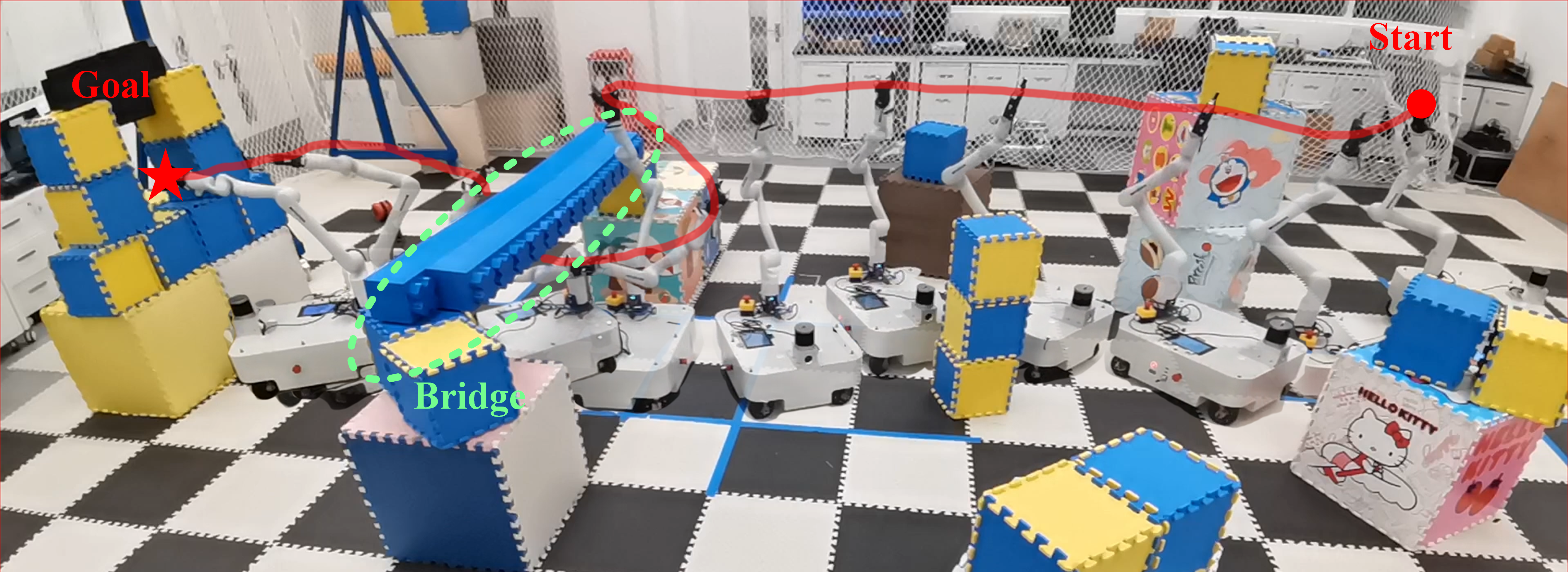}
	\setlength{\abovecaptionskip}{-0pt} 
	\caption
	{Visualization of the overall trajectory in the real-world experiment. The robot is reactive to its environment, avoiding obstacles and passing through a bridge while tracking the pose of its end-effector. The red point indicates the starting position of the end-effector, while the red star marks the target position of the end-effector. }
	\label{fig:illu_real_exp}
\end{figure*}

\subsection{Real-World Experiment}
In this section, we validate our approach in two scenarios. First,  we introduce dynamic obstacles to evaluate the system's capability for dynamic obstacle avoidance.
The location of dynamic obstacles is captured by a motion capture system, and a virtual point cloud of obstacle information is generated at those locations and fed into the algorithm to simulate real obstacles.
Next, we deploy our controller on a mobile manipulator in a $5 \, \text{m} \times 9 \, \text{m}$ cluttered indoor environment, as shown in Fig. \ref{fig:illu_real_exp}. In this experiment, 
the robot is tasked with following a global path of the end-effector while ensuring full-body obstacle avoidance. The motion capture system is employed to determine the state of the mobile base, whereas the manipulator’s state is obtained through the integrated encoder. The point cloud is pre-constructed using the simultaneous localization and mapping‌ (SLAM) system. Meanwhile, we only use a limited perception range as the input of our method to imitate local perception conditions.
Overall, the real-world experiment verifies the robust performance of our approach in navigating clutter and narrow environments.

\section{CONCLUSION}
In this paper, we present a novel approach for controlling mobile manipulators in complex environments. Inspired by time-domain trajectory optimization, we develop a real-time, whole-body safe reactive controller by transforming the single-step control problem into a spatial trajectory optimization problem. 
Our method combines precise link-specific geometric models with extracted free regions around each link, leading to a more accurate and effective formulation for collision avoidance constraints. 
By leveraging the propagation of kinematic chains, our approach preserves sparsity and provides a new routine to accelerate computation through link decoupling. 
We solve this spatial trajectory optimization problem using a DDP-based method integrated with SOS subproblem. 
Extensive simulations and real-world experiments
 demonstrate that our algorithm ensures safe and stable motion for mobile manipulators with specific geometries in cluttered environments. This signifies a substantial advancement in enhancing safety, efficiency, and adaptability of mobile manipulators in challenging scenarios.

\bibliographystyle{IEEEtran}
\normalem
\bibliography{ref}

\end{document}